\documentclass[10pt]{article}

\usepackage{genclaw-techreport}

\usepackage[numbers, sort&compress, square]{natbib}
\usepackage{xspace}
\usepackage{adjustbox}
\usepackage{array}
\usepackage{float}
\usepackage{dblfloatfix}
\usepackage{setspace}
\usepackage{footmisc}
\usepackage{booktabs}
\usepackage{multirow}
\usepackage{tabularx}
\usepackage{threeparttable}
\usepackage{siunitx}
\usepackage{mathtools}
\usepackage{bm}
\usepackage{dsfont}
\usepackage{subcaption}
\usepackage{tikz}
\usepackage{listings}
\usepackage{cleveref}
\usepackage{fontawesome5}
\usepackage{CJKutf8}
\usepackage{arydshln}
\usepackage{multicol}
\usepackage{mdframed}
\usepackage{wrapfig}
\usepackage[ruled,vlined]{algorithm2e}
\usepackage{nicefrac}
\usepackage{fancyvrb}
\usepackage{fvextra}
\usepackage{sourcecodepro}

\definecolor{nipsbg}{gray}{0.98}
\definecolor{nipsframe}{gray}{0.2}

\newtcolorbox{nipsbox}[1]{
    enhanced,
    breakable,
    sharp corners,
    boxrule=0.5pt,
    colframe=nipsframe,
    colback=nipsbg,
    fonttitle=\bfseries\sffamily\small,
    coltitle=white,
    colbacktitle=nipsframe,
    attach boxed title to top left={yshift=-2mm, xshift=3mm},
    boxed title style={sharp corners, boxrule=0pt, top=0.5mm, bottom=0.5mm, left=2mm, right=2mm},
    title={\texttt{#1}},
    top=4mm, bottom=2mm, left=3mm, right=3mm
}

\title{GenClaw: Code-Driven Agentic Image Generation}

\scaiauthor{1,2,3}{Junyan Ye}
\scaiauthor{2}{Jun He}
\scaiauthor{2}{Zilong Huang}
\scaiauthor{4}{Dongzhi Jiang}
\scaiauthor{1}{Xuan Yang}
\scaiauthor{1,†}{Rui Chen}
\scaiauthor{3,†}{Weijia Li}

\affiliation{1}{Tencent Hunyuan}
\affiliation{2}{Sun Yat-sen University}
\affiliation{3}{Tsinghua University}
\affiliation{4}{The Chinese University of Hong Kong}

\metadata[GitHub]{\url{https://github.com/yejy53/GenClaw}}

\begin{abstract}
Image generation models have evolved from text-conditioned pixel synthesis toward multimodal agents endowed with visual comprehension and tool invocation capabilities. Yet, existing agents remain at the mercy of underlying black-box image models. Their workflow is trapped in a repetitive cycle of prompt rewriting for generation refinement, leaving them with no mechanism to directly manipulate the canvas. In essence, the potential of LLMs to serve as a genuine "brush" for precise visual construction remains largely untapped.

In this paper, we propose \textsc{GenClaw}, a code-driven agentic image generation paradigm that empowers the agent to create like a human artist: first conceptualizing, then sketching, and finally coloring.
Specifically, the agent first constructs the conceptual knowledge and context through search and reasoning.
It then utilizes code (e.g., SVG, HTML, Three.js) to render executable visual sketches.
Finally, it employs an image generation model to supplement textures, materials, and photorealism.
In this workflow, code serves as a controllable intermediate canvas bridging linguistic reasoning and pixel synthesis, seamlessly integrating programmatic logic with the visual expressiveness of generative models.
By transforming image generation from a black-box paradigm into a staged process akin to authentic human creation, \textsc{GenClaw} offers a step toward for highly controllable and interpretable visual generation systems.
\end{abstract}

\begin{document}
\maketitle

\begin{figure*}[h!]
    \centering
    \vspace{-2mm}
    \includegraphics[width=\linewidth]{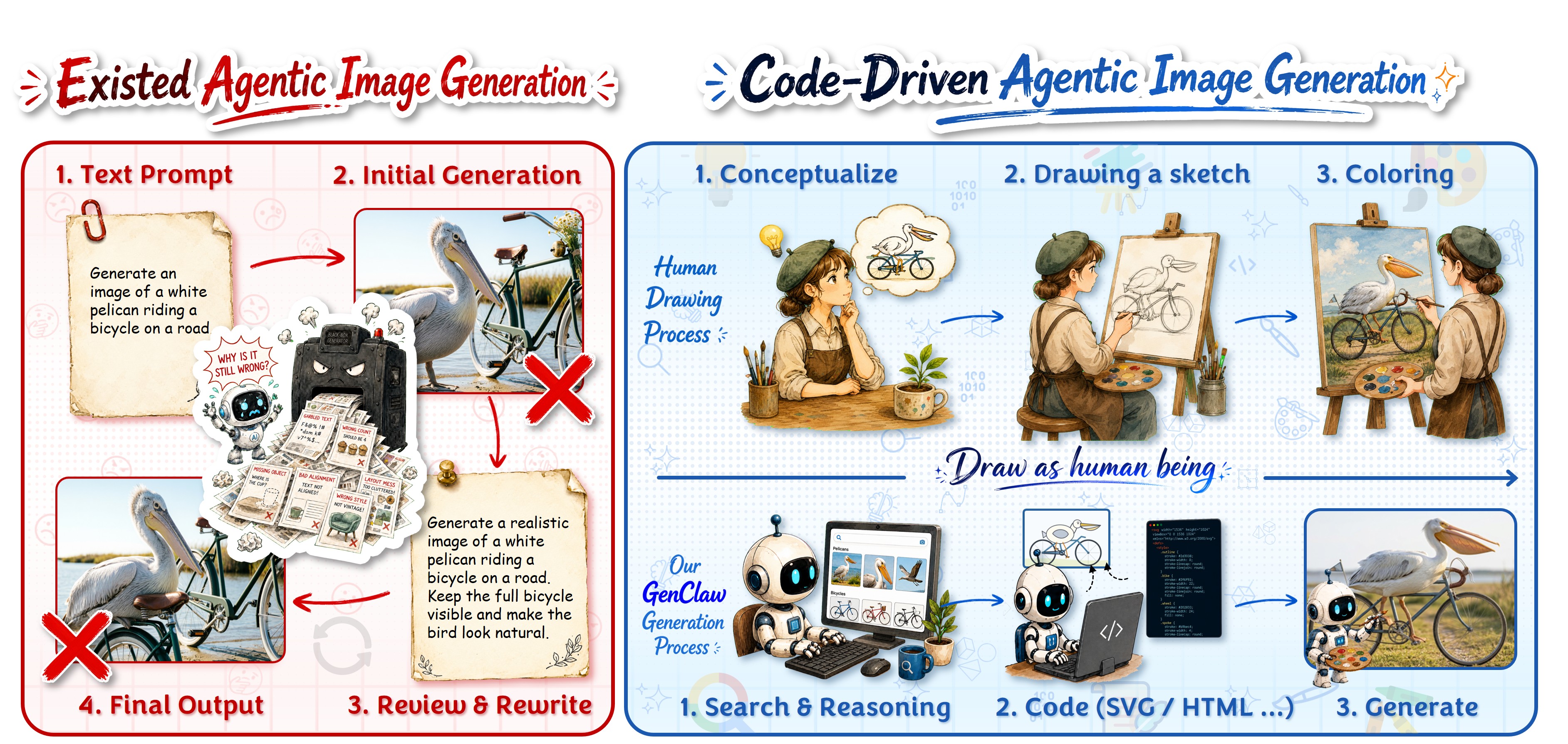}
    \caption{\textbf{Code-Driven Agentic Image Generation.} Existing methods are bottlenecked by end-to-end black-box pixel generators, relying solely on prompt modification for repeated trial-and-error and stochastic sampling.In contrast, GenClaw mimics human creation (conceptualize → sketch → color) by decoupling comprehension from generation. It leverages search and reasoning for context, employs code as a "paintbrush" for precise layout planning, and finally renders the visual output.}
    \label{fig:teaser}
\end{figure*}

\section{Introduction}

Image generation has witnessed remarkable breakthroughs in recent years, with its underlying paradigm steadily transitioning from early text-conditioned synthesis~\citep{LDM,dalle2,zhang2023adding} to unified architectures that seamlessly integrate visual understanding and generation~\citep{chen2025janus,deng2025bagel,wu2025omnigen2,chen2025blip3}. Early GANs and diffusion models~\citep{goodfellow2020generative,podell2023sdxl,sd35Large} significantly propelled the advancement of high-quality pixel synthesis. However, these models serve primarily as text-to-image ``translators,'' exhibiting limited capabilities in deeply comprehending user intent and handling complex logical reasoning. As research progresses, unified understanding-generation models---such as GPT-Image~\citep{openai2024gptimage1}, Qwen-Image~\citep{wu2025qwen}, and Nano-Banana~\citep{team2025gemini}---have elevated the field to unprecedented heights. Driven by massive scaling in model capacity and training data, these large multimodal models demonstrate exceptional capabilities on highly challenging tasks, including world knowledge incorporation, complex instruction following, and typographic text rendering, thereby laying the foundation for next-generation visual generation systems.

In recent advances, image generation is no longer confined to one-shot, end-to-end pixel synthesis. The role of generative models is transitioning from ``passive pixel responders'' to ``Generation Agents'' capable of autonomous planning, tool invocation, and continuous refinement based on feedback~\citep{jiang2026genagent,ye2025mindbrush}. Proprietary systems such as Nano-Banana Pro~\citep{deepmind2024geminiimage}, FLUX 2 Pro~\citep{blackforestlabs2026flux2pro}, and GPT-Image 2~\citep{openai2026gptimage2} have begun integrating Search and Review functionalities, exhibiting a clear trend toward evolving into ``creative agents.'' In academia and the open-source community, works like Think-Then-Generate~\citep{kou2026think} and GenAgent~\citep{jiang2026genagent} explicitly decouple high-level comprehension from concrete generation. Furthermore, JarvisEvo~\citep{lin2025jarvisevo} and RefineEdit-Agent~\citep{liang2025llm} construct closed-loop editing frameworks through the synergy of multimodal CoT and evaluators. Along this trajectory, CoCo~\citep{li2026coco}---while not a fully-fledged agent---generates structured sketches via code prior to refinement, exploring the potential of executable programs as intermediate representations. Notably, Mind-Brush~\citep{ye2025mindbrush} introduces search and reasoning tools into generation, utilizing an agentic architecture to address generative models' deficits in real-time knowledge and complex logic. Concurrently, commercial creative platforms like Lovart\footnote{\url{https://www.lovart.ai/}} and TapNow\footnote{\url{https://www.tapnow.ai/}} are driving the interface paradigm shift from a solitary prompt box to multi-tool collaboration.

However, an in-depth analysis of existing image generation agents reveals a fundamental limitation: although agents play a crucial role in context completion and result review, the final visual synthesis relies almost entirely on end-to-end text-to-image generation. As illustrated in Figure~\ref{fig:teaser}, the agent acts merely as a client giving orders to a printing press, restricted to a stochastic "black-box lottery" via continuous prompt rewriting. Ultimately, this reduces the agent to a glorified "advanced prompt optimizer." In contrast, authentic artistic creation is a highly transparent and staged workflow: human artists wield a paintbrush to seamlessly progress from conceptualization and spatial planning to sketching, and finally to coloring and detailing. In the current agentic paradigm, however, the internal information flow relies almost exclusively on natural language. Inherently, natural language suffers from severe ambiguity when articulating absolute spatial coordinates, exact object counts, complex typographical layouts, and layer occlusion relationships. Consequently, agents fail to acquire substantive operational control over visual-spatial structures. The root cause of this bottleneck is clear: existing agents lack a genuine "paintbrush" tailored to their own modality expertise.

\begin{figure*}[!t]
    \vspace{-1.1cm}
    \centering
    \includegraphics[width=0.99\linewidth]{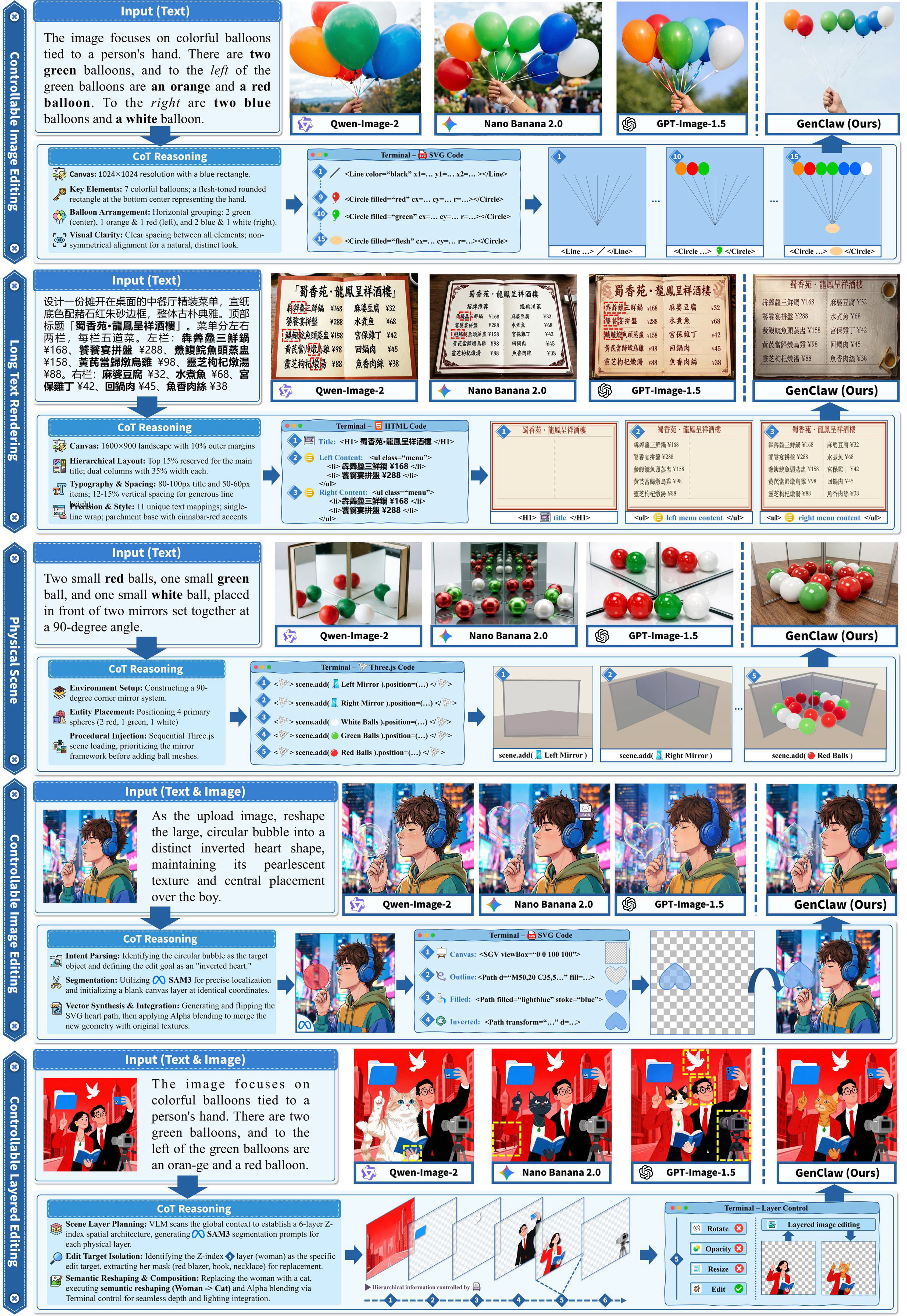}
    \caption{Showcase of GenClaw in complex scene composition, text rendering, physical simulation, and layered image editing.}
    \label{fig:intro_cases}
    \vspace{-1.9cm}
\end{figure*}

To alleviate the limitations of natural language in spatial expression, we need to explore a new kind of ``digital brush'' for agents: an intermediate representation that has better controllability and is natively suited to LLM. In recent years, visual code generation and layered representations have gradually entered the research field and become potential candidates for structured visual generation~\citep{yang2026omnisvg,wang2025internsvg,lin2025vcode,yin2025qwen}. Different from black-box pixel synthesis, code, such as vector programs like SVG, naturally has the advantages of explicit structure, logical rigor, editability, and renderable verification, which fits the programming and debugging capabilities of code agents. In fact, frontier large language models have already shown remarkable potential in code drawing and front-end rendering, allowing them to try to build the skeleton of an image through code, much like a painter sketching line art. However, the strength of code lies in ``logic and structure,'' not in ``pixels and texture.'' If pure code alone is used to render the final image, the result often remains at the level of flat icons, UI, or other regular tasks. This is because pure code has clear expressive bottlenecks when representing high-frequency realistic details such as complex lighting, feathered edges, hair, and natural textures. Realistic image generation is precisely the domain in which image generation models are better.

Based precisely on this natural complementarity in capabilities, this paper proposes a new code-driven agentic image generation paradigm and builds a concrete agent system, \textsc{GenClaw}, based on it, as shown in the right half of Figure~\ref{fig:teaser}. The image generation agent begins to truly imitate the creative role of a painter: it first obtains accurate entity knowledge and context through search and reasoning (\emph{Conceptualize}); then it uses code writing as the ``digital brush'' in its hand to structurally express visual intent on the canvas (\emph{Sketch}), planning object positions, sizes, text layout, layer occlusion ($z$-order), and even 3D physical rules; the image generation model, meanwhile, focuses more on acting as a ``colorist.'' It no longer needs to completely ``blindly guess'' the image structure, but instead colors the structured code sketch generated by the agent (\emph{Color}), supplementing the high-fidelity textures, materials, and realism required by the image. The preliminary system shown in this technical report demonstrates the potential of this decoupled architecture on multiple complex visual tasks that traditional black-box models find difficult to handle stably, as shown in Figure~\ref{fig:intro_cases}:

\begin{itemize}
\item \textbf{More controllable composition.} By compiling complex instructions into visual code with coordinate and quantity references, this paradigm alleviates, to a certain extent, the hallucination problems of traditional models in object counts and spatial relations, and improves the stability of compositional generation tasks.
\item \textbf{More reliable text layout.} Returning text rendering to code, such as SVG or HTML, reduces spelling confusion caused by traditional models treating text as pixel texture fitting, and allows the agent to control font size, alignment, and hierarchy at a finer granularity.
\item \textbf{Assisted simulation of physical laws.} When facing complex physical environments, the system can try to call HTML or Three.js to preliminarily construct a 3D scene with lighting and perspective references, use deterministic computation to assist the expression of physical laws.
\item \textbf{Structured visual-condition editing.} By converting natural language into structured visual code, the agent can more directly manipulate the visual condition input of the underlying generation model, reducing the model's burden of understanding complex language instructions.
\item \textbf{More flexible layered image editing.} By invoking specialized tools, the agent decomposes the image into discrete layers organized via a structured JSONL format. During localized editing, this representation allows the agent to precisely isolate target layers, significantly mitigating unintended pixel corruption in unmodified regions.
\end{itemize}

Ultimately, the genuine paradigm shift is not merely a transition from simple to more complex Prompt Engineering. Instead, it represents a more profound leap: shifting from end-to-end black-box generation to "draw like a human artist." GenClaw's generative workflow aligns closely with the authentic human creative process, thereby exhibiting significant advantages in generation transparency. For instance, upon a generation failure, we can precisely trace the root cause: whether it originates from erroneous context retrieved during search, a logical anomaly when the LLM generates the code-based sketch, or a visual discrepancy during the final sketch-to-photorealistic rendering. This essentially realizes a relatively transparent and traceable pipeline across the entire creative process—from conceptualization and sketching to the final output.

Furthermore, as code agents such as Claude Code and Codex demonstrate extraordinary generalization capabilities and versatile utility, a natural question arises: how can code agents be utilized for visual generation? While previous image generation models operated predominantly as passive chatboxes, the future will inevitably pivot toward an agentic paradigm. GenClaw takes an exploratory step in this direction, serving as an initial harness for image generation. Through GenClaw, we explore how the next generation of image generation agents can achieve highly controllable and interpretable visual synthesis.

\section{Related Work}
\label{sec:related}

\subsection{Image Generation Models}
\label{sec:related_gen}

In recent years, image generation has evolved from text-conditioned
pixel synthesis toward unified large multimodal models that natively
support both visual understanding and generation~\citep{jiang2025draco, jiang2025t2i}. Early diffusion systems (such as Stable Diffusion~\citep{LDM} and DALL-E~\citep{dalle2}) have significantly propelled the rapid advancement of high-quality image synthesis~\citep{ye2025realgen}, demonstrating remarkable performance across diverse image generation tasks~\cite{ye2025leveraging, ye2025echo, li2024crossviewdiff}. Current generative models can synthesize highly photorealistic images that are virtually indistinguishable to the human eye~\citep{cai2025z,ye2025realgen,ye2025loki,wen2026spot}. Subsequent work, most
notably Janus~\citep{chen2025janus}, began to model visual understanding and generation
jointly within a single framework, signaling a gradual shift of the
research focus from single-purpose generators toward more complete
multimodal systems. GPT-4o~\citep{gpt4o} further expanded this trajectory, drawing
attention not only for its generation quality but also for its
complex visual reasoning, text rendering, and instruction-following
abilities. Building on this foundation, follow-up work has deepened
the exploration of architectures and task coverage: BAGEL~\citep{bagel} employs a
Mixture-of-Transformers to separate understanding and generation
experts within a unified architecture and to inject explicit reasoning
into the generation process; Qwen-Image~\citep{wu2025qwen} and its successors perform
well on complex typography and bilingual Chinese/English text
rendering, showing that unified models can scale to more demanding
structured-vision tasks; 
and Nano-Banana~\citep{team2025gemini} achieves solid performance on complex generation and
high-fidelity editing.

Further progress has begun to push image generation models toward an
agentic form. Closed-source systems such as Nano-Banana-Pro~\citep{google2025gemini3} and
FLUX~2 Pro~\citep{blackforestlabs2026flux2pro} have started to integrate search and review modules into
the generation loop, reflecting a visible trend of visual generators
evolving from passive synthesizers into tool-using agents. Taken
together, this trajectory---from single-purpose pixel synthesizers,
to unified understanding-and-generation models, to agentic image
generators---broadens the task boundary of generative models and
provides our work with a strong visual-decoding substrate on which
the code-as-brush paradigm can be built.

\subsection{Agents for Image Generation}
\label{sec:related_agent}

As the capabilities of large language models have grown, the rise of
code agents such as Codex and Claude~\citep{claude} Code suggests that these models
are evolving from conversational assistants into \emph{executable
agents} that read state, invoke tools, and revise their actions
based on feedback. This trend has spawned parallel agentic
approaches for image generation\citep{feng2026gensearch, chen2026unifyagent, ren2026scope}. Think-Then-Generate and GenAgent~\citep{jiang2026genagent}
explicitly decouple high-level understanding from concrete
generation, inserting a multimodal reasoning step before synthesis.
Mind-Brush~\citep{ye2025mindbrush} incorporates search and reasoning tools into open-domain
creation to bridge real-time knowledge gaps. JarvisEvo and
RefineEdit-Agent build closed-loop editing frameworks via multimodal
chain-of-thought and editor--evaluator coordination, supporting
multi-round visual feedback. Commercial systems such as Lovart and
TapNow similarly move creative interfaces away from a single prompt
box toward multi-tool workflows.

Closest to our work, CoCo~\citep{li2026coco} explores using Matplotlib code to produce a
structured sketch that is subsequently refined into a final image,
providing an initial validation of executable programs as an
intermediate representation. However, CoCo still relies heavily on a
single unified model to perform both code generation and pixel
refinement, and therefore does not fully exploit the benefits of a
decoupled architecture on complex tasks. More broadly, existing
image generation agents tend to act as sophisticated prompt
optimizers or knowledge retrievers, with internal information flow
still routed primarily through natural language. As a result,
language models retain limited operational control over visual
spatial structure. In contrast, our code-driven agentic paradigm
materializes the intermediate representation as executable visual
code, allowing the language model to participate directly in
composition, typography, and layered construction, while the image
generation model, acting as the visual decoder, specializes in final
texture expression and photorealistic rendering.

\subsection{Visual Code Generation and Layered Representations}
\label{sec:related_code}

Motivated by the native strengths of large language models in
logical reasoning and code authoring, the use of visual code and
layered representations to guide image generation has emerged as an
active research direction~\citep{yang2026omnisvg,wang2025internsvg,lin2025vcode}. Unlike direct pixel-space synthesis,
these approaches represent visual content as vector programs composed
of paths, shapes, text, and hierarchies (e.g., SVG, HTML), which are
editable, losslessly scalable, and structurally explicit. OmniSVG~\citep{yang2026omnisvg} is
the first to model high-quality SVG generation as a unified
multimodal task, demonstrating end-to-end capability from simple
icons to complex illustrations. InternSVG~\citep{wang2025internsvg} further integrates SVG
understanding, editing, and generation within the same framework,
exploring vector code as a shared intermediate language across
tasks. As foundation models grow stronger, general-purpose language
models exhibit non-trivial potential for zero-shot code-based
drawing: Kimi k2.5~\citep{team2026kimi} and DeepSeek~V4~\citep{deepseekai2026deepseekv4} both demonstrate the ability to
construct complex physical structures or render web interfaces
directly from code, suggesting that writing visual code is becoming a
native skill of frontier language models. In parallel, VCode~\citep{lin2025vcode} shows
that SVG can serve as an intermediate representation for
visual-semantic compression and revision; Vec2Pix~\citep{guo2026controlling} demonstrates that
hierarchical SVG can act as a bridge toward high-fidelity pixel
images; and Qwen-Image-Layered~\citep{yin2025qwen} and related layered-representation
work argue that explicitly decomposing an image's structure is a
meaningful path toward more editable visual models.

However, existing pure-code generation research is largely confined
to relatively regular tasks such as icons, UI layouts, and isolated
components; its ability to support overall composition of complex
scenes or open-domain semantic organization remains limited.
Moreover, pure code has inherent expressive limits when rendering
high-frequency, photorealistic details such as lighting, hair, and
natural texture. Motivated by this observation, we do not treat
visual code as a final product; instead, we treat it as a
code-based intermediate sketch inside the agent, used to decompose
the image, organize the layout, and support iterative revision,
while the final photorealistic rendering is delegated to the image
generation model acting as a visual decoder.

\section{Method}
\label{sec:method}

\begin{figure*}[!t]
    \centering
    \includegraphics[width=\linewidth]{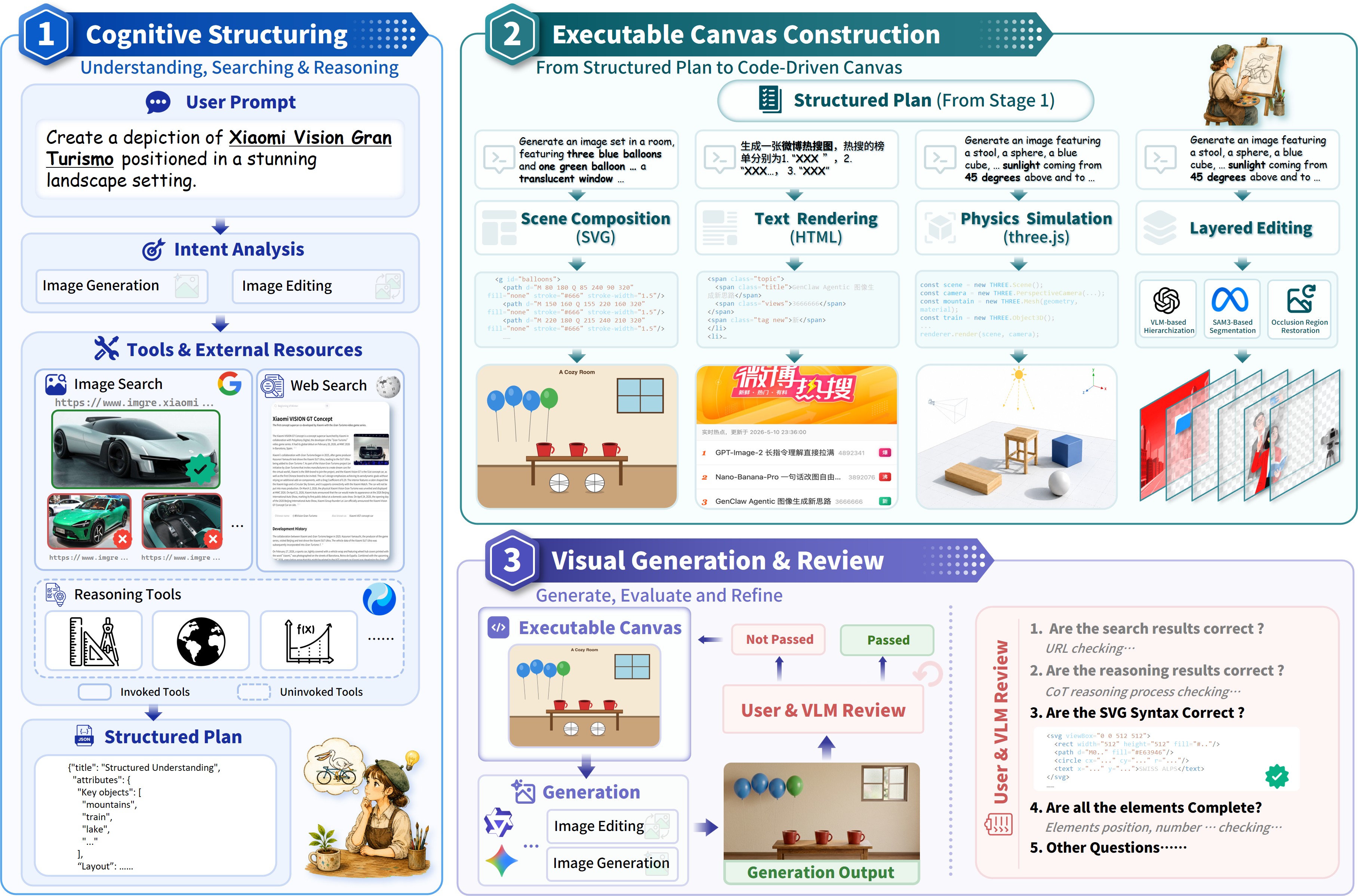}
    \caption{\textbf{Overall architecture of the proposed framework.} By emulating the human drawing workflow, the agentic pipeline is decoupled into three corresponding layers: (1) Cognitive Structuring Layer (Think) for intent understanding, context search, and complex reasoning; (2) Executable Canvas Layer (Sketch), which uses code as a "digital paintbrush" to construct precise intermediate layouts; and (3) Visual Generation and Review Layer (Color) for final image rendering and VLM-based validation.}
    \label{fig:pipeline}
\end{figure*}

\subsection{Overall Framework}
\label{sec:method-overall}

This paper proposes a code-driven image generation agent framework. Its core idea is to turn image generation from a black-box process that directly sends a prompt to an model into a staged process that is closer to how humans draw. When humans create an image, they usually do not obtain the final picture at the beginning. Instead, they first form an idea in mind and, when necessary, search for references or reason about the task; then they make a draft to determine the objects, positions, text, and main structure; finally, they add textures, lighting, and realism. In the agent system, we decompose this process into three layers: cognitive structuring, executable canvas construction, and visual generation and review.

As shown in Figure~\ref{fig:pipeline}, the first layer is the \textbf{Cognitive Structuring Layer}. In this layer, the agent uses a VLM/LLM as the core cognitive module, together with search, knowledge bases, and reasoning tools, to actively complete the understanding work before image generation. This includes understanding the user's intent, understanding reference images, completing world knowledge, and performing mathematical, geographic, and physical reasoning. These tasks are cognitive activities that multimodal agents are good at, rather than the main responsibility that should be carried by an image generation model. The second layer is the \textbf{Executable Canvas Layer}. In this layer, the agent converts the structured records organized by the first layer into an executable canvas, such as SVG, HTML/CSS, Python plotting, or a simple 3D script. Code here acts as the agent's ``digital brush'': instead of asking the agent to draw through GUI mouse operations, we let the agent directly construct objects, text, coordinates, layers, and editable units through CLI/code forms that better match its own capabilities. The third layer is the \textbf{Visual Generation and Review Layer}. The agent invokes off-the-shelf image generation models (e.g., Qwen-Image, Nano Banana) to render the intermediate executable canvas into visually rich final images. Subsequently, the synthesized results are reviewed—either automatically by a VLM or interactively by the user—to ensure precise alignment with the target objectives. Owing to the inherent transparency of the agentic workflow, users are empowered to perform highly dynamic, fine-grained content adjustments and interactions based on both the intermediate layouts and the final outputs.

Compared with a one-step prompt-to-image approach, our framework makes the agent's thinking no longer stay only at rewriting prompts, but further materializes it as an executable canvas state. The final image is not completely ``extracted'' by the image model from the prompt. Instead, similar to a human white-box creation process, the agent first thinks and conceptualizes, then builds a sketch, and finally completes image creation.

\subsection{Cognitive Structuring Layer}
\label{sec:method-cognitive}

The goal of this layer is to decouple the understanding and reasoning tasks before image generation from the image generation model, and to establish a cognitive trajectory between the user's intent and the executable canvas. This corresponds to the thinking and conception stage in human drawing. Existing generation models are good at mapping conditions to visual content, but they are not always suitable for complex intent parsing, world-knowledge retrieval, or symbolic reasoning. Different from directly expanding a prompt, this layer does not try to directly generate the final image description. Instead, it explicitly parses the user's intent, completes missing knowledge, or performs necessary reasoning analysis, and organizes the results as structured records.

Specifically, the agent first performs intent understanding. For ordinary image generation tasks, the agent may only conduct lightweight prompt organization. However, when the task involves specific or dynamic concepts, such as long-tail entities, real-time events, geographic locations, cultural symbols, or professional objects, the model's internal knowledge is often insufficient to support accurate generation. At this time, the agent calls search tools to complete the relevant facts, thereby filling the cognitive gap. For requests involving mathematics, geography, physics, and other tasks that require complex understanding and reasoning, the agent first explicitly obtains intermediate conclusions based on the VLM's reasoning ability, and then converts implicit relations into visual constraints. For instance, in geometry tasks, the agent computes the numerical answer prior to rendering it visually. This process makes the cognitive work before image generation explicit, instead of leaving all understanding pressure to the final image model.

After completing intent understanding, knowledge completion, and reasoning analysis, the agent organizes the results into JSONL-style structured records. Unlike natural-language prompts, these records do not pursue descriptive richness, but emphasize executability and traceability: they need to specify which objects should appear, which text should be rendered, which relations should be preserved, and which knowledge facts support these visual decisions. Such structured records are also useful for interaction between human users and the agent. For example, in drawing a science-popularization poster, the user can explicitly query whether the knowledge content is wrong.

\subsection{Executable Canvas Layer}
\label{sec:method-canvas}

After completing the conception of the drawing content, the agent constructs sketch content in a way similar to human drawing. The agent selects an appropriate programming backend according to the task type, and compiles the objects, text, spatial relations, and constraints into executable code. After execution, the code generates a sketch-like intermediate image, which carries the core layout, text content, and structural relations of the image.

For complex compositional scene generation, such as tasks that emphasize quantity and spatial relations, the agent can use SVG as the canvas backend. SVG can explicitly represent each object as a node and control layout through coordinates, size. For example, in scenes that require generating a fixed number of objects, strict left-right relations, or occlusion relations, the agent can directly create the corresponding number of object nodes in SVG, rather than relying on the image model to understand natural-language descriptions such as ``how many'', or ``located at the center''.

For text-intensive tasks, the agent can use HTML/CSS to build the canvas. Menus, course schedules, webpage cards, instruction pages, and infographics often contain large amounts of Chinese, English, prices, titles, and section information. At this time, the agent deterministically draws the text content through a renderer, and then hands it to the subsequent image model for visual enhancement, avoiding asking the image model to directly ``guess-write'' text as pixel texture. In addition, in some tasks where the text does not need to be re-rendered and regenerated, we can also adopt a strategy where the text is directly rendered by code and the background image is generated by the image model.

For tasks involving physical laws, the agent can call Python plotting, Canvas, or lightweight 2D/3D code to build geometric references. The key to this type of task is not visual style, but correct relations. For example, based on Three.js~\citep{threejs} code, the system can explicitly place a mirror and render the reflection result of a small ball, as shown in Figure~\ref{fig:teaser}. At this time, code plays the role of a physical simulator, deterministically modeling the mirror reflection law before final image generation.

For editing tasks, the agent tends to first build a layered representation of the image. The agent first understands the image content based on a VLM, then divides the objects in each layer, and then uses SAM~\citep{sam} 3 tools for object segmentation. It also uses an image model to complete occluded regions. After layered representation, the agent can treat the contents as editable layers and control transparency and rendering order through a JSONL format. For example, when the user asks to move an object, replace a piece of text, or modify only a certain region in the image, the agent can first locate the corresponding object layer or mask region, and modify its position, content, or attributes.

Traditional image generation agents usually can only indirectly influence the image model by modifying the prompt, and the final structure still depends on the model's sampling result. In contrast, code provides the agent with a more natural operation interface, and it is highly matched with the agent's capability form. The agent is better at reading and writing structured text, calling tools, modifying local code, and checking execution results, rather than drawing stroke by stroke through a mouse in a GUI like a human. Therefore, we do not ask the agent to simulate human GUI operations, but let it directly construct the image structure through CLI/code. Object count, text content, spatial position, and layer relations can all be clearly expressed in code.

\begin{table*}[t!]
\centering
\caption{Quantitative Comparison of different methods on \textbf{GenEval++}~\cite{ye2025echo}. The best performing model within each subgroup (open-source / agentic) is highlighted in \textbf{bold}. }
\label{tab:geneval_performance}
\fontsize{9pt}{12pt}\selectfont
\setlength{\tabcolsep}{1.8mm}
\begin{tabular}{lcccccccc}
\toprule
Method & Count & Color & Color/Count & Color/Pos & Pos/Count & Pos/Size & Multi-Count & Overall \\
\midrule
\multicolumn{9}{c}{\textbf{Open-source Models}} \\
\midrule
SD-3.5 M~\cite{esser2024scaling} & 0.600 & 0.400 & 0.250 & 0.250 & 0.075 & 0.400 & 0.300 & 0.325 \\
FLUX.1-dev~\cite{flux2024} & 0.600 & 0.400 & 0.250 & 0.250 & 0.075 & 0.400 & 0.300 & 0.325 \\
Janus Pro 7B~\cite{chen2025janus} & 0.300 & 0.450 & 0.125 & 0.300 & 0.075 & 0.350 & 0.125 & 0.246 \\
T2I-R1~\cite{jiang2025t2i} & 0.325 & 0.675 & 0.200 & 0.350 & 0.075 & 0.250 & 0.300 & 0.311 \\
BLIP3-o~\cite{chen2025blip3} & 0.325 & 0.675 & 0.200 & 0.350 & 0.075 & 0.250 & 0.300 & 0.311 \\
OmniGen2~\cite{wu2025omnigen2} & 0.325 & 0.675 & 0.200 & 0.350 & 0.075 & 0.250 & 0.300 & 0.311 \\
Bagel~\cite{deng2025bagel} & 0.600 & 0.325 & 0.250 & 0.325 & 0.250 & 0.475 & 0.375 & 0.371 \\
DraCo~\cite{jiang2025draco} & 0.530 & 0.450 & 0.280 & 0.400 & 0.280 & 0.530 & 0.380 & 0.400 \\
Qwen-Image~\cite{wu2025qwen} & \textbf{0.725} & \textbf{0.875} & \textbf{0.725} & \textbf{0.600} & \textbf{0.475} & \textbf{0.725} & \textbf{0.550} & \textbf{0.668} \\
\midrule
\multicolumn{9}{c}{\textbf{Closed-source Models}} \\
\midrule
GPT-Image-1~\cite{openai2024gptimage1} & 0.675 & 0.900 & 0.725 & 0.625 & 0.600 & \textbf{0.800} & \textbf{0.850} & 0.739 \\
Nano-Banana~\cite{team2025gemini} & 0.700 & 0.800 & \textbf{0.680} & 0.530 & \textbf{0.680} & 0.780 & 0.800 & 0.710 \\
GPT-Image-1.5~\cite{openai2025gptimage15} & \textbf{0.950} & \textbf{0.850} & 0.775 & 0.500 & 0.600 & 0.775 & 0.800 & 0.750 \\
Gemini-3.1 Flash-Image~\cite{gemini-2.0-flash} & 0.850 & 0.850 & \textbf{0.800} & \textbf{0.750} & 0.675 & 0.725 & 0.775 & \textbf{0.775} \\

\midrule
\multicolumn{9}{c}{\textbf{Agentic generation Models}} \\
\midrule
PromptEnhancer~\cite{wang2025promptenhancer} & 0.625 & 0.500 & 0.225 & 0.375 & 0.125 & 0.450 & 0.375 & 0.382 \\
GenAgent~\cite{jiang2026genagent} & 0.775 & 0.775 & 0.650 & 0.800 & 0.600 & 0.725 & 0.750 & 0.725 \\
Gemini-3.0 Pro-Image~\cite{google2025gemini3} & 0.900 & 0.800 & 0.750 & 0.700 & 0.725 & 0.725 & 0.850 & 0.761 \\
Mind-Brush~\cite{ye2025mindbrush} & 0.700 & 0.775 & 0.775 & 0.750 & 0.850 & 0.775 & 0.850 & 0.782 \\

\rowcolor{blue!10}GenClaw & \textbf{0.950} & \textbf{0.875} & \textbf{0.825} & \textbf{0.850} & \textbf{0.925} & \textbf{0.825} & \textbf{0.900} & \textbf{0.878} \\

\bottomrule
\end{tabular}
\end{table*}

\subsection{Visual Generation and Review Layer}
\label{sec:method-visual-review}

The third layer is where the agent calls existing image generation or editing models to complete the final visual realization, and uses a VLM or multimodal evaluation ability to review the result. Based on the executable code and its rendered sketch obtained from the second layer, the agent uses it as visual-condition input, and calls Qwen-Image,  Nano Banana~\citep{team2025gemini}, or other models with image generation and editing capabilities to complete the final generation. At this point, the image model no longer needs to plan the scene structure from scratch, but supplements texture, lighting, material, and realism on the basis of the existing sketch and structural constraints.

In generation scenarios, the image model more often plays the role of visual realizer: it performs naturalized rendering based on the code sketch and text condition, concentrating model capability on texture, lighting, material, details, and overall realism, rather than simultaneously undertaking tasks such as complex planning, counting, text layout, or physical reasoning. On the other hand, compared with relying only on an LLM to generate visual code, the image model also breaks through the upper limit of code expression. Code sketches are good at expressing structure, layout, and text, but are difficult to use for complex natural textures, realistic lighting, and open-scene details; the image generation model can preserve the sketch structure while extending visual content from simple UI or schematic diagrams to more natural and realistic scene generation.

In editing scenarios, the agent performs modifications based on editable objects, layered representations, or local masks provided by the second layer. For example, in tasks such as ``move a cup'', ``replace the title in a poster'', or ``change the color of an object'', the agent can first determine the object layer or region that needs to be modified, and then call an image editing model to complete the local visual update. Since the editing target and scope have already been given by the executable canvas, the model does not need to understand the entire image again, and therefore has an advantage in local consistency and preservation of non-target regions.

After generation, the agent's Review module verifies whether the final output aligns with the user's objectives and the initial structured records. In conventional end-to-end image generation, although a VLM or user can ultimately detect synthesis errors, this black-box observation lacks fine-grained interpretability, making it difficult to pinpoint the root cause. In contrast, GenClaw's workflow boasts inherent transparency. The VLM can trace and diagnose issues by inspecting the intermediate representations recorded throughout the entire pipeline. For instance, if an error stems from flawed cognitive comprehension or inaccurate external knowledge, the agent can trace back to the first layer to verify the accuracy of the URL contents retrieved by the search tool. For issues such as incorrect object counts or missing text, the system can cross-reference the intermediate code outputs with the final rendered image to precisely localize the failure. Through this stratified tracing mechanism, the agent effectively decouples cognitive, structural, and visual errors, allocating them to their respective layers for targeted resolution.

\begin{figure*}[t!]
    \centering
    \includegraphics[width=\linewidth]{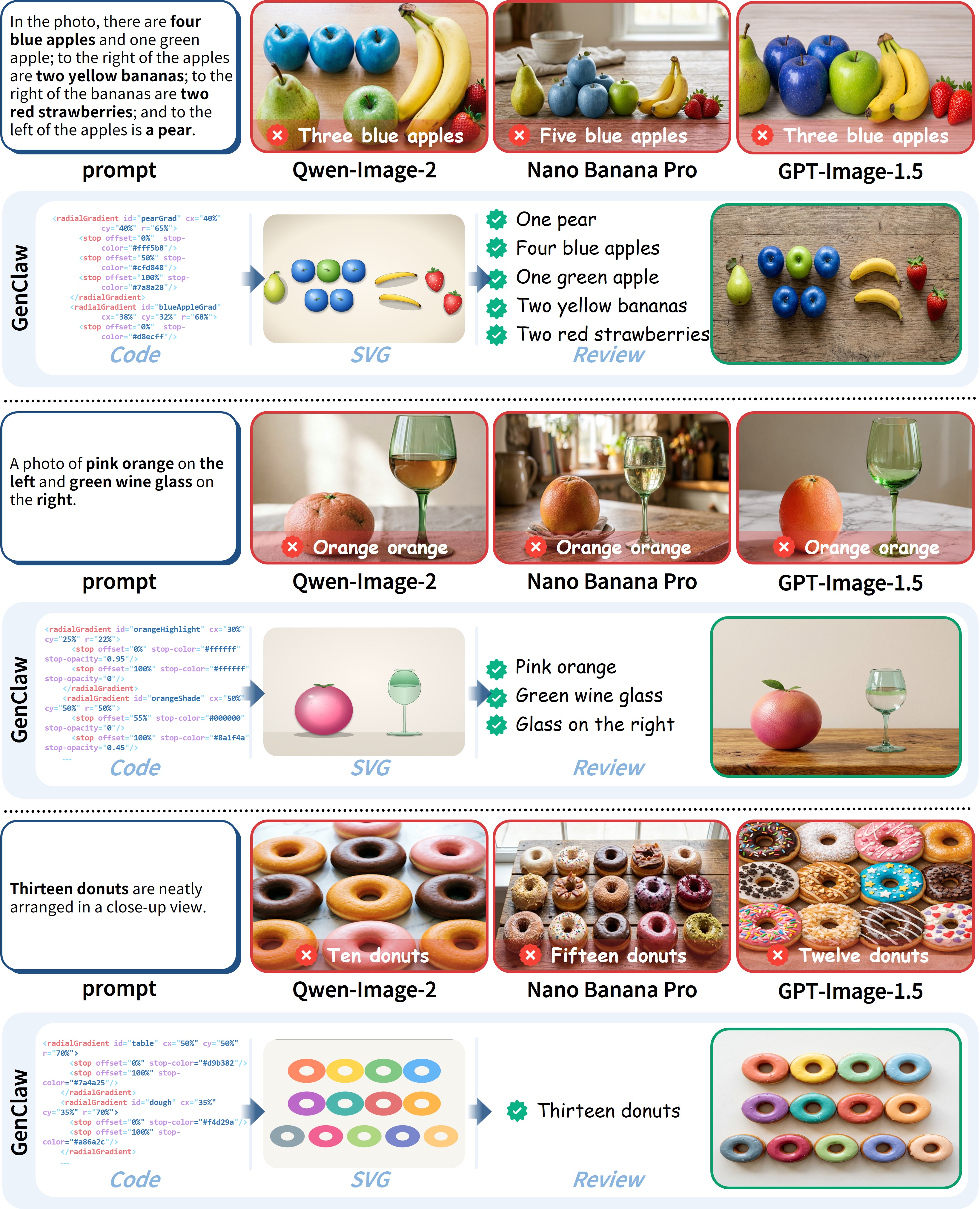}
    \caption{\textbf{Qualitative comparison of instruction following in complex compositions.} Compared to purely text-driven traditional generation, GenClaw leverages LLMs to generate SVG code for explicit layout planning, demonstrating superior performance in complex object counting and multi-attribute binding tasks.}
    \label{fig:ComplexScene}
\end{figure*}

\section{Experiments}
\label{sec:exp}

\subsection{Experimental Setup}
\label{sec:exp_setup}

We evaluate \textsc{GenClaw} across a diverse set of image generation and editing tasks. The selected benchmarks include GenEval++~\citep{ye2025echo} for complex scene instruction following, LongText-Bench~\citep{geng2025x} for text rendering, ImgEdit~\citep{ye2025imgedit} for image editing, and Mind-Bench~\citep{ye2025mindbrush} for assessing world knowledge and reasoning capabilities. For baseline comparisons, we evaluate \textsc{GenClaw} against state-of-the-art open-source and proprietary generative models, including GPT-Image~\citep{openai2024gptimage1}, Qwen-Image~\citep{wu2025qwen}, Nano-Banana~\citep{deepmind2024geminiimage25}, and BAGEL~\citep{bagel}. Furthermore, to explicitly distinguish our "code-as-brush" mechanism from conventional agentic prompt rewriting, we also incorporate image agent systems dominated by the rewriting paradigm, such as GenAgent and Mind-Brush.

Regarding implementation, the agent backbone of \textsc{GenClaw} employs \textbf{Claude-Opus-4.6}, with the default generator set to \textbf{Gemini-3.1-Flash-Image}~\citep{deepmind2024geminiimage25}. The agent is responsible for translating user intents into structured records and executable canvases, while the generator performs the final natural rendering conditioned on sketches, text layers, localized masks, or specific editing constraints. The backend rendering code dynamically adapts to the task at hand: SVG is utilized for structured composition and layer-wise editing; HTML/CSS or SVG text layers are applied to poster design and long-text tasks; and Python, Canvas, or Three.js scripts are adopted for physical and geometric previews.

\subsection{Main Results}
\label{sec:exp_main}

\subsubsection{Executable Structure Improves Compositional Control}
\label{sec:exp_composition}

As shown in the results of Table~\ref{tab:geneval_performance}, on the GenEval++ task, which evaluates instruction following for complex scene layout, \textsc{GenClaw} benefits from the agent's explicit SVG pattern guidance and achieves clear performance advantages on tasks such as Counting and Spatial. It still has an advantage compared with the generation results of GPT-Image-1.5 or Gemini-3.0 Pro-Image. This is because even powerful closed-source models still face challenges when they rely only on the controllability of text, especially for tasks where natural-language descriptions of quantity and space are easily compressed or mismatched.

This result also distinguishes \textsc{GenClaw} from rewrite-prompt paradigms such as Mind-Brush, GenAgent, and PromptEnhancer. Rewrite-based agents can improve the instruction-following ability of the base model by expanding and reorganizing prompts, and therefore usually improve over direct generation. However, their intermediate state is still natural language and cannot truly lock object counts and spatial coordinates. For tasks that require exact attribute binding, the agent still can only repeatedly modify prompts and resample. In contrast, \textsc{GenClaw} writes these discrete structures into an SVG canvas: object nodes, positions, sizes, and layers are already explicitly determined before generation, and the visual decoder only needs to supplement texture, and realism on this structure.

Figure~\ref{fig:ComplexScene} provides the corresponding qualitative evidence. For instructions containing multiple objects, attributes, and spatial relations, direct generators can usually produce visually reasonable images, but they easily suffer from problems such as count errors and attribute-binding failures. The intermediate sketch of \textsc{GenClaw} is closer to the draft stage in human drawing: it first uses a code-based sketch to clarify the image skeleton, and then enters naturalized rendering. Therefore, the improvement in Table~\ref{tab:geneval_performance} does not come from a ``longer prompt'', but from a structural sketch drawn by the agent based on code and executable checking.

\subsubsection{Text Rendering and Poster Generation}
\label{sec:exp_text}

Table~\ref{tab:longtext} shows the text rendering ability of \textsc{GenClaw} on LongText-Bench. Compared with existing image generation models, \textsc{GenClaw} achieves clear advantages on both Chinese and English long-text tasks. This result comes from a change in task division: text is no longer ``guessed'' by the image model in pixel space, but is deterministically rendered by HTML/SVG text layers. The image model is mainly responsible for background, style, and visual details, and therefore does not need to simultaneously undertake character generation, layout organization, and realistic rendering. This performance advantage is particularly pronounced in fine-grained generation scenarios such as HTML pages, slides, and posters, where the code representation intrinsically possesses strong expressive power.

\begin{figure*}[t!]
    \centering
    \includegraphics[width=0.97\linewidth]{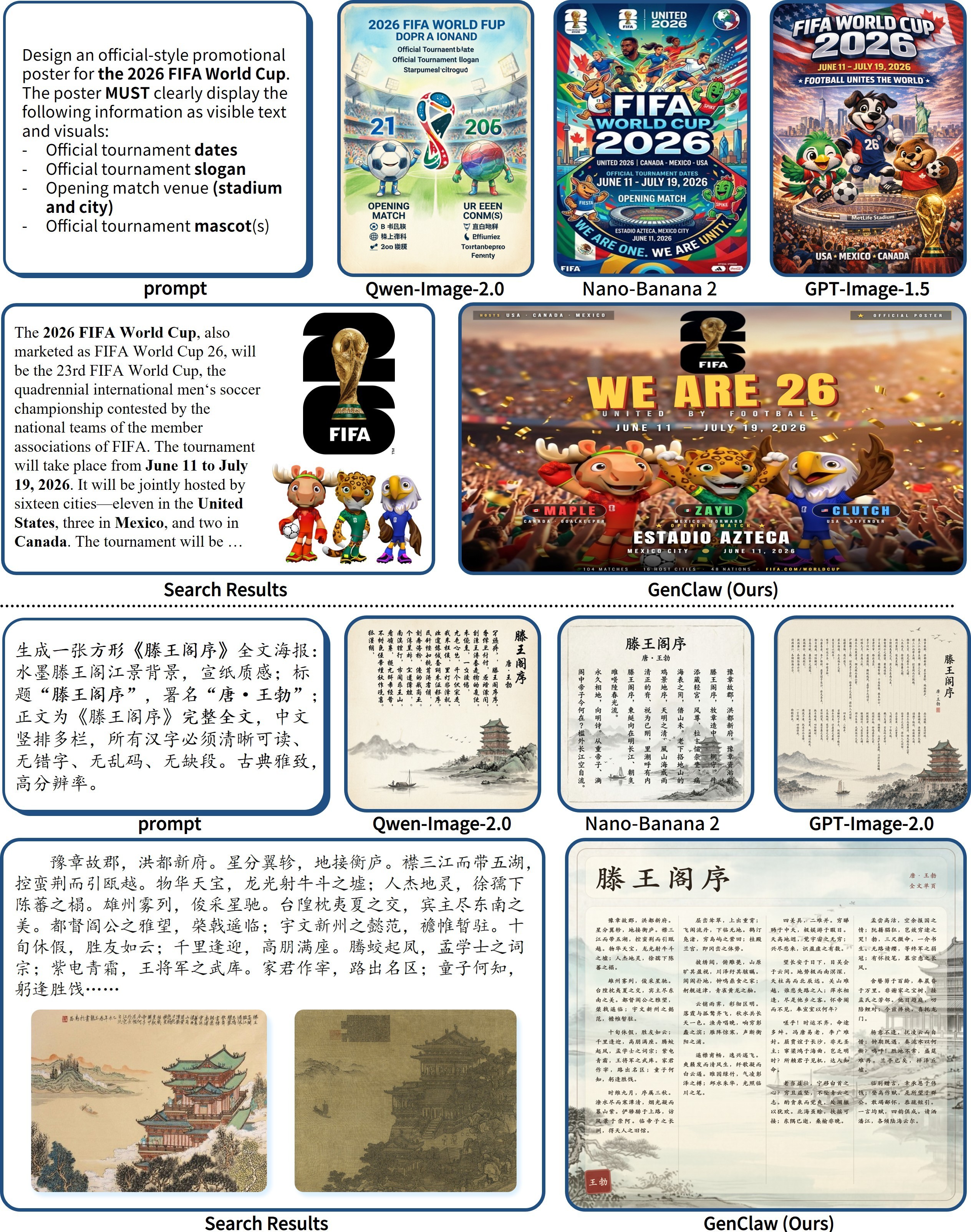}
    \caption{\textbf{Qualitative comparison of long-text poster design.} GenClaw demonstrates strong tool-use synergy: it retrieves world knowledge via the search tool to fill in missing context, while leveraging a code-driven engine for precise typographic layout and rendering, ensuring highly accurate text generation.}
    \label{fig:TEXT}
\end{figure*}

\begin{table}[t]
\centering
\caption{Quantitative comparison on \textbf{LongText-Bench}. We report the official metrics for both English and Chinese long-text rendering.}
\label{tab:longtext}
\begin{tabular}{l c c}
\toprule
\textbf{Model} & \textbf{LongText-Bench-EN} $\uparrow$ & \textbf{LongText-Bench-ZH} $\uparrow$ \\
\midrule
FLUX.1-dev~\cite{flux2024} & 0.607 & 0.005 \\
OmniGen2~\cite{wu2025omnigen2} & 0.561 & 0.059 \\
BAGEL~\cite{deng2025bagel} & 0.373 & 0.310 \\
Lumina-DiMOO~\cite{qin2025lumina-image2} & 0.437 & 0.047 \\
InternVL-U~\cite{chen2024internvl} & 0.738 & 0.860 \\
X-Omni~\cite{geng2025x} & 0.900 & 0.814 \\
LongCat-Next~\cite{team2026longcat} & 0.932 & 0.891 \\
Z-Image~\cite{cai2025z} & 0.935 & 0.936 \\
Qwen-Image~\cite{wu2025qwen} & 0.943 & 0.946 \\
Emu3.5~\cite{cui2025emu3} & 0.976 & 0.928 \\

GPT-Image-1~\cite{openai2024gptimage1} & 0.956 & 0.619 \\
Seedream 4.5~\cite{seedream42025} & 0.989 & 0.987 \\
Gemini-3.0 Pro-Image~\cite{deepmind2024geminiimage} & 0.981 & 0.949 \\

\midrule
\textbf{GenClaw} & 0.989 & 0.988 \\
\bottomrule
\end{tabular}
\end{table}

The poster-making cases in Figure~\ref{fig:TEXT} further illustrate the relation between text rendering and world-knowledge completion. Taking examples such as the ``2026 World Cup'' or ``Preface to the Pavilion of Prince Teng'', the model not only needs to generate accurate text, but also needs to know what information should be presented: the former involves real-time events, hosting information, visual symbols, and layout organization, while the latter involves classical text content, cultural context, and typographic aesthetics. \textsc{GenClaw} can call search tools to complete relevant knowledge, organize the retrieved facts into structured content, and then write them into an HTML/SVG canvas. The final image generation model only needs to undertake background image, decorative element, and overall style generation, instead of handing knowledge organization, text layout, and character drawing all to the same black-box pixel model. In addition, the content knowledge retrieved by the agent can also be directly understood or edited by the user, greatly improving the interpretability of the image generation process.

\subsubsection{Physical Simulation as Executable Visual Reasoning}
\label{sec:exp_physics}

\begin{figure*}[t!]
    \centering
    \includegraphics[width=1\linewidth]{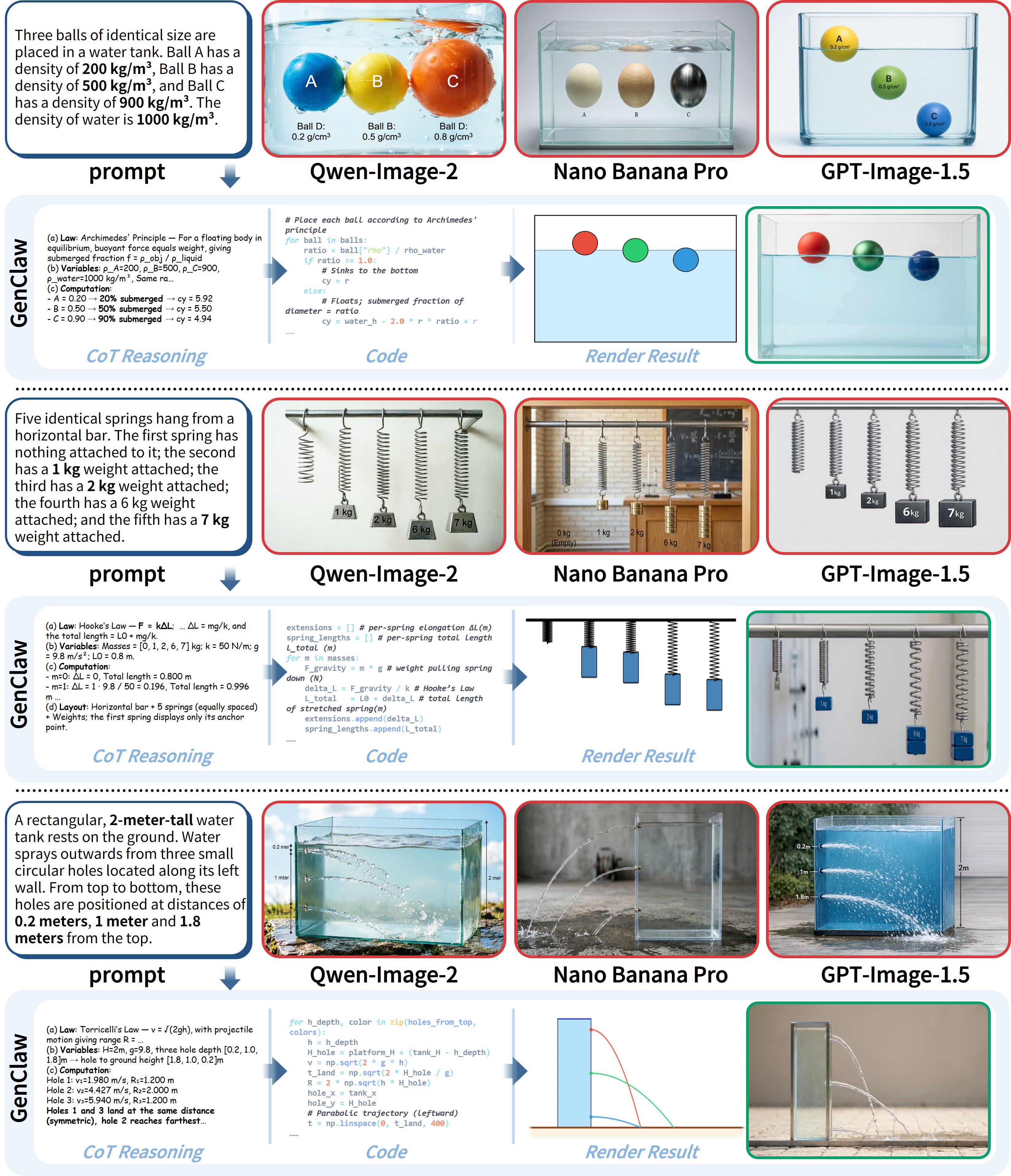}
    \caption{\textbf{Image generation governed by physical laws.} To overcome the inherent deficits of visual models in intuitive physics, GenClaw decouples comprehension from generation. It first executes code for specific physical simulations (e.g., spring deformation, water jet range) to derive precise metrics, which then drive the image rendering.}
    \label{fig:Phy}
\end{figure*}

Figure~\ref{fig:Phy} shows our preliminary exploration of physical-simulation image generation tasks with \textsc{GenClaw}. Different from ordinary composition or text rendering, these tasks more strongly test whether an image generation model can understand and present compound physical laws. From the results, whether it is the mirror rendering problem in Figure~\ref{fig:Phy}, or specific physical scenes such as springs, pressure, and buoyancy, direct image generation models do not perform ideally. The reason is that image models are often better at fitting the object content mentioned in the text, but do not necessarily understand what physical relations these elements should satisfy.

In contrast, before executing image generation, \textsc{GenClaw} first uses code to build a simplified physical or geometric model. At this time, code is not only a visual layout tool, but also plays the role of intermediate modeling. For example, in the mirror problem, the system can first place mirror material, light sources, and object positions based on Three.js~\citep{threejs}, and perform deterministic rendering, so that final image generation can refer to the reflection position in the sketch. For pressure, springs, and other problems, the agent also first parses physical variables, constraint relations, and visualization goals from the user request, and then uses Python, Canvas, and other tools to build an intermediate simulation image. This intermediate image does not pursue final realism, but serves as a ``physical draft'' or ``symbolic world model'', first ensuring that structures such as reflection, spring deformation, force direction, liquid level, or geometric relations are correct, and then handing it to the visual decoder for naturalized rendering.

This attempt shows that the value of code-as-brush is not limited to ``drawing more neatly''. When a generation task contains formalizable world rules, the code canvas can execute part of world modeling before pixel generation, converting implicit physical relations into inspectable visual constraints. Using code as an intermediate representation has the potential to push visual generation from ``imagining the world based on text'' toward ``first building an executable simplified world, then performing visual rendering''.

\begin{table}[b]
\centering
\caption{Quantitative comparison on \textbf{ImgEdit}. We utilize the VLM-Score to measure the efficacy of editing instruction execution, while employing PSNR and SSIM to evaluate the image consistency within unedited regions.}
\label{tab:imgedit}
\begin{tabular}{l c c c}
\toprule
\textbf{Method} & \textbf{VLM-Score} $\uparrow$ & \textbf{PSNR} $\uparrow$ & \textbf{SSIM} $\uparrow$ \\
\midrule
FLUX.1 Kontext~\cite{fluxkontext}         & 3.43 & 19.59 & 0.592 \\
Qwen-Image-Edit~\cite{wu2025qwen}         & 3.70 & 17.64 & 0.526 \\
Uniworld-V2 (Qwen-Image-Edit)~\cite{uniworldv2} & 3.72 & 16.13 & 0.482 \\
CoCoEdit~\cite{cocoedit}        & 3.79 & 19.13 & 0.555 \\
FLUX.2 Klein 9B~\cite{flux2klein2026}     & 4.33 & 22.50 & 0.695 \\

Gemini-3.1 Flash~\cite{gemini-2.0-flash}  & 4.51 & 20.19 & 0.583 \\
GPT-Image-1.5~\cite{openai2025gptimage15} & 4.69 & 16.36 & 0.433 \\
\midrule
\textbf{Ours}                           & 4.29 & 27.87 & 0.718 \\
\bottomrule
\end{tabular}
\end{table}

\subsubsection{Image Editing on ImgEdit}
\label{sec:exp_imgedit}

\begin{table*}[t!]
    \centering
    \caption{Quantitative comparison of different models on \textbf{Mind-Bench}. The table is divided into two sections: conventional generative models (top) and agentic generative models (bottom). The best-performing results are highlighted in \textbf{bold}. The symbol "—" indicates that the model is not applicable to Image-to-Image (I2I) tasks.}
    \label{tab:mindbench}
    \fontsize{8pt}{9pt}\selectfont
    \setlength{\tabcolsep}{1.9mm}
    \resizebox{\textwidth}{!}{%
    \begin{tabular}{l|ccccc|ccccc|c}
        \toprule
        \multirow{2}{*}{\textbf{Model Name}} & \multicolumn{5}{c|}{\textbf{Knowledge-Driven}} & \multicolumn{5}{c|}{\textbf{Reasoning-Driven}} & \multirow{2}{*}{\textbf{Overall}} \\
        \cmidrule(lr){2-6} \cmidrule(lr){7-11}
             & SE & Weather & MC & IP & WK & SL & Poem & Life Reason & GU & Math & \\
            \midrule

            FLUX 1 dev~\citep{flux2024} & 0.04 & 0.00 & 0.00 & 0.00 & 0.02 & 0.02 & 0.04 & - & - & - & 0.02 \\
            FLUX 1 Kontext~\citep{labs2025flux1kontextflowmatching} & 0.02 & 0.00 & 0.00 & 0.00 & 0.02 & 0.00 & 0.00 & - & - & - & 0.01 \\
            BAGEL~\citep{deng2025bagel} & 0.02 & 0.00 & 0.00 & 0.00 & 0.00 & 0.02 & 0.02 & 0.02 & 0.00 & 0.08 & 0.02 \\
            Z-Image~\citep{cai2025z} & 0.02 & 0.00 & 0.08 & 0.02 & 0.00 & 0.00 & 0.00 & - & - & - & 0.02 \\
            Qwen-Image~\citep{wu2025qwen} & 0.08 & 0.00 & 0.04 & 0.00 & 0.00 & 0.04 & 0.00 & 0.04 & 0.00 & 0.00 & 0.02 \\
            GPT-Image-1~\citep{openai2024gptimage1} & 0.32 & 0.06 & 0.22 & 0.02 & 0.16 & 0.32 & 0.10 & 0.24 & 0.10 & 0.12 & 0.17 \\
            GPT-Image-1.5~\citep{openai2025gptimage15} & 0.36 & 0.18 & 0.22 & 0.04 & 0.30 & 0.34 & 0.08 & 0.34 & 0.10 & 0.02 & 0.21 \\
            FLUX 2 Max~\citep{blackforestlabs2026flux2max} & 0.26 & 0.34 & 0.02 & 0.00 & 0.34 & 0.32 & 0.52 & 0.20 & 0.18 & 0.10 & 0.23 \\
            Nano Banana~\citep{deepmind2024geminiimage25} & 0.24 & 0.20 & 0.12 & 0.00 & 0.36 & 0.32 & 0.40 & 0.28 & 0.08 & 0.24 & 0.22 \\
           
            \midrule

            FLUX 2 Pro~\citep{blackforestlabs2026flux2pro} & 0.28 & 0.32 & 0.02 & 0.00 & 0.20 & 0.36 & 0.58 & 0.20 & 0.16 & 0.12 & 0.22 \\
            Nano Banana Pro~\citep{deepmind2024geminiimage} & 0.52 & 0.24 & 0.20 & 0.04 & 0.52 & 0.60 & 0.72 & 0.28 & \textbf{0.44} & 0.28 & 0.38 \\
            Mind-Brush~\citep{ye2025mindbrush} & 0.54 & 0.16 & 0.62 & 0.18 & 0.40 & 0.26 & 0.54 & 0.10 & 0.16 & 0.14 & 0.31 \\
            \rowcolor{blue!20} \textsc{GenClaw} & \textbf{0.64} & \textbf{0.44} & \textbf{0.66} & \textbf{0.32} & \textbf{0.64} & \textbf{0.78} & \textbf{0.90} & \textbf{0.38} & 0.32 & \textbf{0.60} & \textbf{0.57} \\
            \bottomrule
    \end{tabular}%
    }
\end{table*}

We evaluate image editing performance on the ImgEdit benchmark. Beyond standard VLM-based holistic scoring, we place particular emphasis on the image consistency of unedited regions. Therefore, following CoCoEdit~\citep{cocoedit}, we utilize mask annotations to compute the PSNR and SSIM exclusively on the unedited areas, with the results summarized in Table~\ref{tab:imgedit}. As observed, most baseline models yield relatively low PSNR and SSIM scores, indicating substantial alterations to the unedited regions. In particular, despite achieving high VLM-based evaluation scores, GPT-Image-1.5 obtains sub-optimal consistency metrics, suggesting that it introduces highly aggressive modifications to non-target areas during the editing process. In contrast, \textsc{GenClaw} demonstrates a substantial improvement in pixel-level preservation metrics. This indicates that \textsc{GenClaw} inflicts significantly less disruptive corruption on non-target areas, an advantage directly attributed to the inherent protection afforded by the layer-wise editing paradigm. Furthermore, while Qwen-Image-Layered is currently geared more towards simple layer separation tasks, it exhibits limited controllability for fine-grained layer-wise editing. Although our current image decomposition mechanism remains relatively rudimentary, we view that layered representation is of paramount value—not only for profound image comprehension but also for the unification of understanding and generation tasks. This remains a focal point for our future explorations.

\begin{figure*}[h]
    \centering
    \includegraphics[width=\linewidth]{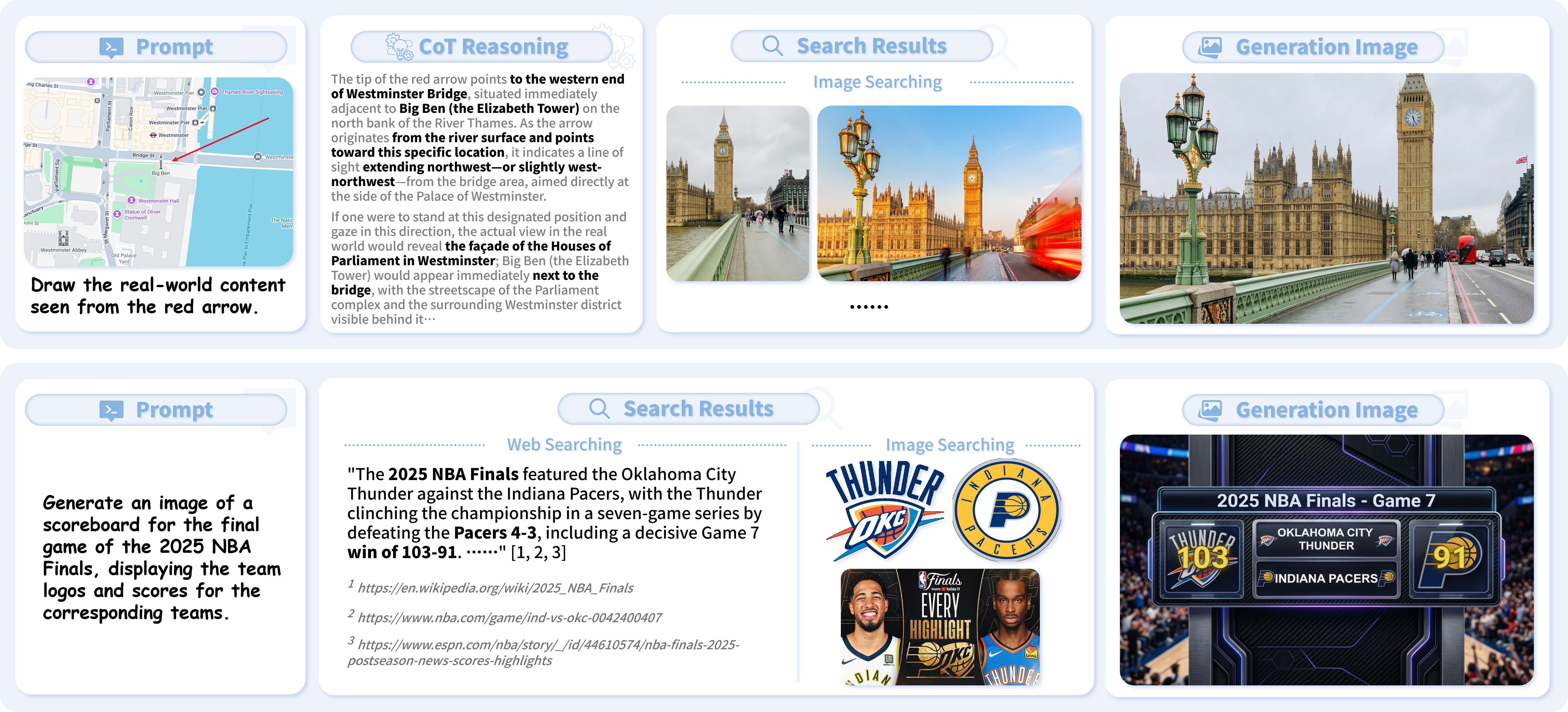}
    \caption{\textbf{Visualization of GenClaw's results on Mind-Bench.} Prior to the final image rendering, the image agent invokes reasoning or search tools to gather sufficient contextual information.}
    \label{fig:MindBench}
    \vspace{-0.7cm}
\end{figure*}

\subsubsection{Knowledge Grounding on Mind-Bench}
\label{sec:exp_mindbench}

Mind-Bench~\citep{ye2025mindbrush} is a benchmark focused on knowledge-driven and reasoning-driven image generation tasks, and can test whether a model correctly understands implicit intent and external facts before generation. The experimental results are shown in Table~\ref{tab:mindbench}. Compared to base image models, image agents—including Mind-Brush and Nano-Banana-Pro—achieve superior performance. Meanwhile, \textsc{GenClaw} continues to deliver highly competitive results. Compared to Mind-Brush, which relies solely on a single-pass Google search, \textsc{GenClaw} significantly optimizes the search workflow for acquiring external knowledge and retrieving relevant images: it incorporates a multi-round search mechanism and filters the most suitable reference samples from multiple retrieved candidates. By adopting this paradigm that decouples comprehension from generation, \textsc{GenClaw} further demonstrates that incorporating agentic external knowledge and explicit reasoning can effectively enhance the performance of image models.

Figure~\ref{fig:MindBench} provides qualitative visualizations of GenClaw on Mind-Bench. For instance, in a street view synthesis task for a specific location, the agent first invokes the reasoning tool to identify and infer the contents of the input map. It then utilizes the web-search tool to retrieve authentic street views of that location as references before proceeding to image generation. Similarly, for tasks requiring deterministic factual knowledge (e.g., precise NBA game scores), the agent employs search tools to ground the specific informational context prior to visual rendering. Furthermore, this mechanism allows users to review the generated results in a relatively "white-box" manner, enabling them to easily trace and verify the factual correctness throughout the entire generation pipeline.

\section{Limitations and Future Work}
\label{sec:limitations}

While the code-driven agentic paradigm, as demonstrated by \textsc{GenClaw}, significantly enhances spatial controllability and compositional accuracy, it also presents several limitations that highlight directions for future research:

\paragraph{High Dependency on the Underlying Generation Model.}
The sketches rendered from visual code (e.g., SVG or HTML) are inherently abstract. Translating these abstract structures into high-fidelity, photorealistic images requires exceptional generalization capabilities from the underlying image generation model. In our experiments, we observed that current open-source conditional generation models often struggle with this task, frequently producing severe artifacts, degrading textures, or simply retaining the flat, original SVG style rather than achieving photorealism. Consequently, to fully validate the feasibility of this decoupled paradigm at the current stage, our research must rely on powerful frontier models like Gemini-3.1-flash Image.

\paragraph{Efficiency Overhead and Diminishing Returns.}
\textsc{GenClaw} introduces a multi-step agentic pipeline, which inevitably incurs significant inference latency and computational overhead. While this time cost is highly justified for complex generation tasks that require precise control, the long pipeline becomes overly redundant and inefficient for simple, straightforward tasks compared to traditional one-shot end-to-end generation. Furthermore, as the native capabilities of foundational image generation models continue to advance, many complex tasks that currently require an agentic workflow may eventually be handled directly by the models themselves. As a result, the marginal gain provided by the agent architecture for image generation might gradually diminish in the future.

\paragraph{Stability Risks in Code Generation.}
The process of translating natural language into executable code carries inherent instability. LLMs are not entirely infallible when generating code; they may occasionally produce errors such as coordinate calculation deviations, incorrect layer occlusion ($z$-order) relationships, or disproportionate element scaling. These code-level flaws directly manifest in the rendered sketches, leading to suboptimal spatial layouts or misaligned details in the final generated images, thereby limiting the system's stability in certain scenarios.

\section{Conclusion}
\label{sec:conclusion}

In summary, the genuine paradigm shift advocated in this work is not from Prompt Engineering toward more sophisticated Prompt Engineering, but rather from ``letting a model guess an image in one shot'' toward ``letting an agent build the skeleton of an image step by step through code, like a human painter.'' Centered on this philosophy, we introduce the Code-Driven Agentic Image Generation paradigm and instantiate it in our system \textsc{GenClaw}: through the \emph{Conceptualize $\rightarrow$ Sketch $\rightarrow$ Color} workflow, the LLM is dedicated to what it excels at---logic and structure---while the image generation model returns to its native strength of pixels and texture. This decoupled design not only demonstrates stronger controllability across complex composition, text rendering, physical simulation, and layered editing, but also renders the entire generation process transparent and traceable: any failure can be localized to a specific stage---retrieval, code generation, or final rendering---an advantage that end-to-end black-box models inherently lack.

Looking ahead, just as code agents like Claude Code and Cursor are profoundly reshaping the software engineering paradigm, we believe the field of image generation is undergoing a similar evolution, progressively merging into this broader agentic wave. While previous generative models predominantly operated in a passive, chatbox-based response mode, future visual generation systems will inevitably pivot toward a more proactive and agentic paradigm. \textsc{GenClaw} represents our initial exploratory step in this direction. We hope our work can inspire the community and provide a valuable reference for building the next generation of visual creation systems endowed with high controllability, interpretability, and profound reasoning capabilities.

\newpage

\bibliographystyle{plainnat}
\bibliography{ref}

\end{document}